# Tacit Knowledge Management with Generative AI: Proposal of the GenAI SECI Model

**Naoshi Uchihira[1]**

[1]Japan Advanced Institute of Science and Technology, Ishikawa, 923-1211, Japan

**ABSTRACT**

The emergence of generative AI is bringing about a significant transformation in knowledge management. Generative AI has the potential to address the limitations of conventional knowledge management systems, and it is increasingly being deployed in real-world settings with promising results. Related research is also expanding rapidly. However, much of this work focuses on research and practice related to the management of explicit knowledge. While fragmentary efforts have been made regarding the management of tacit knowledge using generative AI, the modeling and systematization that handle both tacit and explicit knowledge in an integrated manner remain insufficient. In this paper, we propose the "GenAI SECI" model as an updated version of the knowledge creation process (SECI) model, redesigned to leverage the capabilities of generative AI. A defining feature of the "GenAI SECI" model is the introduction of "Digital Fragmented Knowledge", a new concept that integrates explicit and tacit knowledge within cyberspace. Furthermore, a concrete system architecture for the proposed model is presented, along with a comparison with prior research models that share a similar problem awareness and objectives.

**Keywords:** Knowledge Management, Tacit Knowledge, SECI Model, Generative AI

## INTRODUCTION

In recent years, the latest AI technologies, including generative AI, have been bringing about major industrial and social transformations — a phenomenon referred to as AI Transformation (AX). For example, business support through conversational AI agents powered by Large Language Models (LLMs), and AI-assisted coding tools such as GitHub Copilot, have the potential not merely to improve operational efficiency, but to fundamentally reshape industrial structures, including the nature of work itself and the way people work.

Generative AI is also driving significant change in the field of knowledge management. Conventional knowledge management systems have had several major challenges, and even when organizations built and deployed such systems and accumulated knowledge within them, the systems themselves often could not be put to effective use. Specifically, organizing and registering knowledge into a system required considerable effort, and users frequently struggled to retrieve the information they needed through search. Generative AI, however, holds the potential to resolve these challenges (O'Leary, 2024), and cases of generative AI being applied to knowledge management are indeed on the rise.

Nevertheless, the application of generative AI to knowledge management has, until recently, been largely confined to explicit knowledge. At the same time, efforts to extract and leverage tacit knowledge from the workplace (referred to as "Gen-Ba" in Japan) using generative AI have been growing in recent years — a trend that is particularly active in Japan. This is due to the following reasons. Workplace tacit knowledge has long been a source of strength and competitive advantage for Japanese companies. However, as Japan faces an accelerating demographic shift toward an aging society with a declining birthrate, there is an increasingly urgent need to efficiently and effectively transfer workplace tacit knowledge from veteran employees to younger employees before the former retire. In practice, pilot efforts to leverage generative AI for the transfer of workplace knowledge have already begun in leading-edge companies in Japan.

While such efforts to extract and leverage workplace tacit knowledge through generative AI are underway, systematic modeling of these efforts is still insufficient. In this paper, we propose the "GenAI SECI" model — a version of the knowledge creation process (SECI) model updated on the premise of generative AI utilization. We also present a concrete system architecture "Digital Knowledge Twin" based on the GenAI SECI model.

The remainder of this paper is organized as follows. Section 2 presents a review of prior research. Section 3 proposes a knowledge management model that incorporates the use of generative AI, then Section 4 illustrates a concrete system architecture based on the proposed model. Section 5 clarifies the characteristics of the proposed model through comparison with prior studies, and Section 6 offers concluding remarks.

## LITERATURE REVIEW

Among knowledge management models, the SECI model proposed by Nonaka & Takeuchi (1995) is widely known. This model divides knowledge into explicit knowledge and tacit knowledge, and posits that organizational knowledge is created through repeated mutual conversion between the two. The knowledge creation mechanism described by this model effectively explained the competitive advantage of Japanese companies at the time. The SECI model consists of four modes of knowledge conversion (Figure 1):

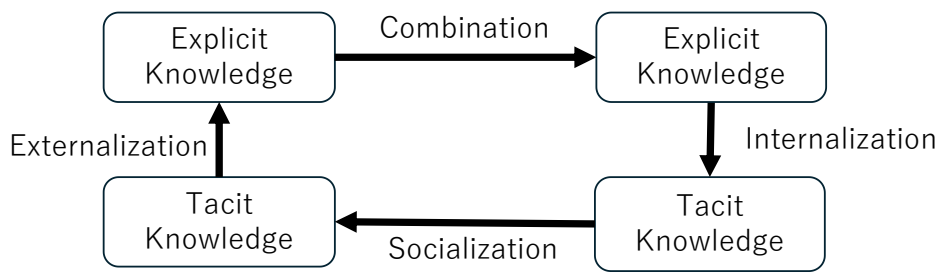

Figure 1: SECI Model. (Adapted from Nonaka & Takeuchi (1995))

•Socialization (Tacit -> Tacit): The process by which tacit knowledge is shared within an organization as tacit knowledge through direct experience sharing, observation, and imitation.

- Externalization (Tacit -> Explicit): The process of converting tacit knowledge into explicit forms such as concepts, metaphors, models, and documents, making experiential knowledge accessible to the entire organization.
- Combination (Explicit -> Explicit): The process of generating new explicit knowledge by combining, categorizing, and integrating multiple forms of explicit knowledge.
- Internalization (Explicit -> Tacit): The process of absorbing explicit knowledge through practice and experience and embodying it as an individual's tacit knowledge.

Because the SECI model is a versatile and easy-to-use framework, various studies based on the SECI model have been conducted in relation to knowledge management utilizing AI including generative AI. These fall into two categories: (1) studies describing the role of AI and generative AI within the four knowledge conversion modes of the SECI model, and (2) studies proposing new models based on the SECI model.

Studies in category (1) include Kai, Purba & Al-Hosaini (2024), Shen & Lin (2024), He & Burger-Helmchen (2025), and Sumbal & Amber (2025). Kai & Purba & Al-Hosaini (2024) is a systematic review of the relationship between AI and knowledge management, pointing out that "AI-driven systems can automate and enhance the processes of externalization and internalization by making tacit and explicit knowledge more accessible and actionable," but does not address generative AI in detail. He & Burger-Helmchen (2025) organizes and discusses how AI can facilitate and accelerate each of the SECI processes (socialization, externalization, combination, and internalization), concluding that the most effective use of AI is to complement and augment rather than replace human capabilities. Shen & Lin (2024) propose an AI-assisted Experience-Based Knowledge Management System (EBKMS), explaining each function of the system in correspondence with each process of the SECI model. Sumbal & Amber (2025) explain how ChatGPT can be used in each process of the SECI model.

Studies in category (2) take the position that the SECI model cannot adequately represent knowledge management utilizing generative AI; specifically, these include Böhm & Durst (2026) and Kirchner & Scarso (2026). Böhm & Durst (2026) propose GRAI (Generative Receptive Artificial Intelligence) as a useful framework for describing the new role of machines (generative AI) in the knowledge creation process. GRAI treats machines (generative AI) as new actors and expands the four interactions of the SECI model by adding human–machine (generative AI) dimensions, extending them to eight interactions. Kirchner & Scarso (2026) also treat generative AI as a new actor, and additionally introduce "Artificial Knowledge"—a new type of knowledge generated by generative AI alongside explicit and tacit knowledge—proposing the AKI (Artificial Knowledge Integration) model. All of these are very recent papers published in 2026, and there remains considerable room for further investigation in this field.

In this study, while taking the position that a new model is needed for knowledge management utilizing generative AI, we propose a new model based on the SECI model (the GenAI SECI model) that differs from Böhm & Durst (2026) and Kirchner & Scarso (2026) in treating generative AI not as a new actor but strictly as an auxiliary means. What is unique and novel about the GenAI SECI model is that it introduces "Digital Fragmented Knowledge" as a new type of knowledge in cyberspace, and utilizes generative AI for the externalization and internalization of this knowledge.

## KNOWLEDGE MANAGEMENT MODEL UTILIZING GENERATIVE AI

In this section, we propose a knowledge management model utilizing generative AI (GenAI SECI model). First, we clarify the knowledge to be managed here — specifically, workplace knowledge ("Gen-Ba" Knowledge (Uchihira et al., 2023)). In Japanese, "Gen" means "actual and physical field," and "Ba" means "knowledge creating space," which was introduced by Nonaka & Takeuchi (1995). Gen-Ba knowledge refers to knowledge that is created, shared, and utilized in the workplace. Gen-Ba includes various settings such as manufacturing sites, maintenance and inspection sites, medical and nursing care settings, agricultural sites, and sales and customer service environments. Furthermore, knowledge here is defined as justified belief that serves as the basis for human action. Here, "justified" means "something that one has genuinely accepted and internalized for oneself." In this paper, Gen-Ba knowledge is classified into three layers: explicit knowledge, latent knowledge, and tacit knowledge (Figure 2). The three layers of knowledge are continuous and do not have clear boundaries between them.

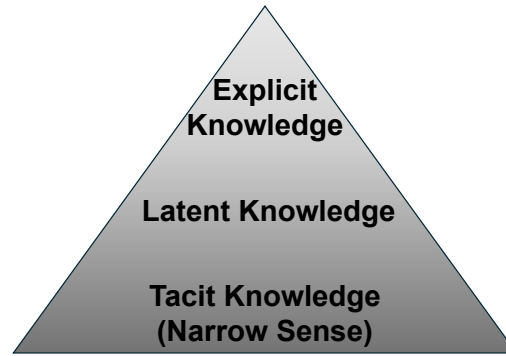

Figure 2: Continuous Gen-Ba Knowledge Layers (Uchihira et al., 2023)

- Explicit Knowledge: Knowledge that is consciously recognized and can be codified (manualized) in a complete and systematic manner.
- Latent Knowledge: Knowledge that is not ordinarily conscious, but can be partially and fragmentarily expressed as code (text, image, video, etc.) when one is in the Gen-Ba or when asked by others.
- Tacit Knowledge (Narrow Sense): Knowledge that exists unconsciously and cannot in principle be expressed in code (verbalized). A representative example of tacit knowledge is embodied knowledge such as the skilled techniques of craftspeople.

The tacit knowledge of the SECI model encompasses both latent knowledge and tacit knowledge (narrow sense), and is referred to here as tacit knowledge (broad sense). Collins (2010) classified tacit knowledge into "Relational Tacit

Knowledge", "Somatic Tacit Knowledge", and "Collective Tacit Knowledge". Somatic Tacit Knowledge corresponds broadly to Tacit Knowledge (Narrow Sense), while Relational Tacit Knowledge corresponds to Latent Knowledge. Regarding Collective Tacit Knowledge, those elements that can be fragmentarily codified are included in Latent Knowledge, while the remainder are classified as Tacit Knowledge (Narrow Sense). It is our view that much of Collective Tacit Knowledge can in fact be partially codified. With respect to Somatic Tacit Knowledge as well, those elements that can be fragmentarily codified may reasonably be included in Latent Knowledge. Here, codification is defined to include meaningful text, images, video, and similar forms of expression.

Furthermore, the proposed model introduces "Digital Fragmented Knowledge". Explicit Knowledge, Latent Knowledge, and Tacit Knowledge (Narrow Sense) are classifications of knowledge from a human perspective. Digital Fragmented Knowledge is defined as knowledge in cyberspace — specifically, Explicit Knowledge and Latent Knowledge that has been accumulated in cyberspace in some digital form. For example, records of what people in the Gen-Ba have sensed or thought, documented in language or photographs, do not constitute systematic knowledge like a manual, but can be accumulated in cyberspace as partial and fragmentary knowledge (referred to as "knowledge fragments"). While such partial and fragmentary knowledge has traditionally been difficult to handle within knowledge management, generative AI has now made it possible to utilize them. Figure 3 shows the proposed GenAI SECI model.

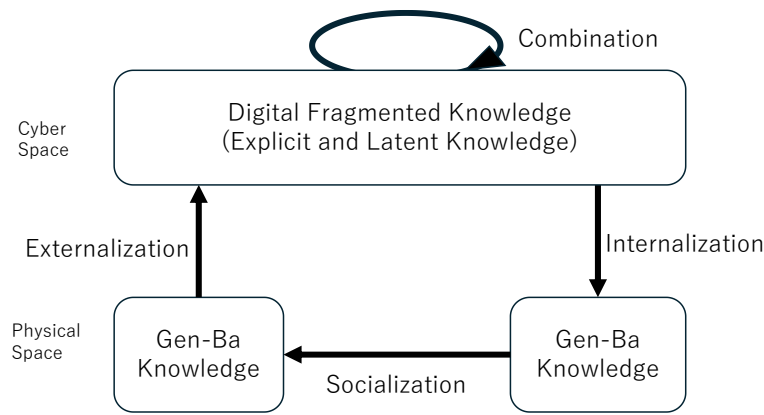

Figure 3: GenAI SECI Model

In this GenAI SECI model, each process is defined as follows.
- Socialization: Same as socialization in the SECI model. The process by which Gen-Ba knowledge is shared within an organization, through direct experience sharing, observation, and imitation.
- Externalization: The process of codifying Gen-Ba knowledge — even partially and fragmentarily — as "knowledge fragments" and accumulating them in cyberspace.
- Combination: The process of organizing the "knowledge fragments" accumulated in cyberspace into a form suitable for internalization.

- Internalization: The process of referencing the "knowledge fragments" accumulated and organized in cyberspace to amplify human Gen-Ba knowledge.

The roles of generative AI in externalization, combination, and internalization of the GenAI SECI model are described below. Specific technologies and functions are provided in the following section (Digital Knowledge Twin System).
- Externalization: When codifying Gen-Ba knowledge distributed across various media (text, sensor data, images, video, etc.) as "knowledge fragments," generative AI aggregates them into meaningful units.
- Combination: Generative AI organizes and stores "knowledge fragments" in a structured format such as a knowledge graph.
- Internalization: Generative AI selects and presents "knowledge fragments" that are effective for amplifying human Gen-Ba knowledge.

The distinctive feature of this model is that, rather than externalizing tacit knowledge into complete explicit knowledge and then using that explicit knowledge for internalization as in the SECI model, it leverages generative AI to amplify Gen-Ba knowledge through internalization using incomplete knowledge fragments, without requiring full externalization into explicit knowledge. A comparison with the GRAI framework and the AKI model from prior studies is provided in the Discussion section.

## DIGITAL KNOWLEDGE TWIN SYSTEM

In this section, as a concrete system architecture based on the GenAI SECI model, we introduce the Digital Knowledge Twin System (Uchihira et al., 2025) shown in Figure 4, which consists of three processes (A), (B) and (C). This system is currently under development.

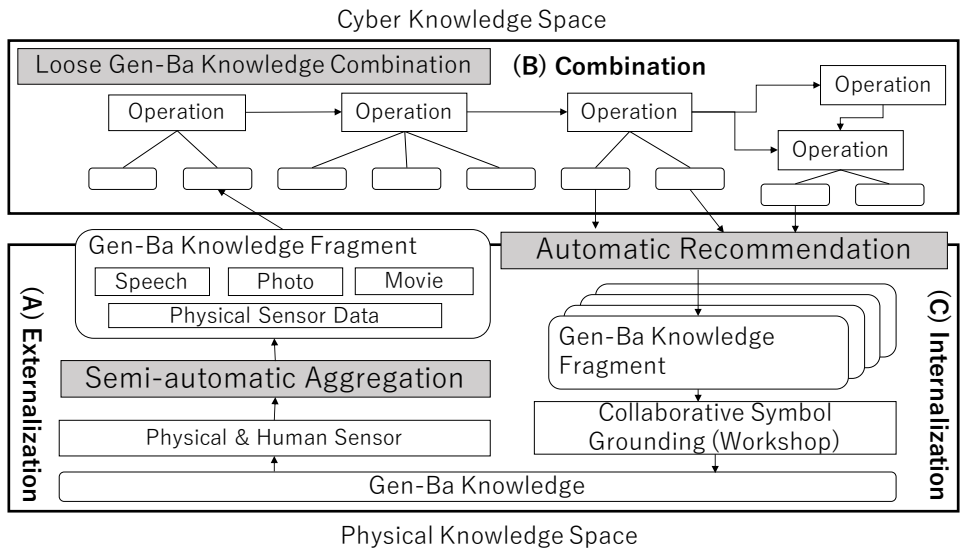

Figure 4: Digital Knowledge Twin System  (Uchihira et al., 2025)

**(A)    Semi-automatic Aggregation in Externalization**

We have extended the Smart Voice Messaging System, developed since 2010 as a human sensor to collect the awareness of skilled workers through voice messages and photos (Uchihira, 2013). This extension allows for the multifaceted recording and accumulation of voice messages, photos, and physical sensor data in the Gen-Ba. In our previous research, the aggregation of voice messages, photos, and physical sensor data (Gen-Ba knowledge fragment) was manually performed by humans, which was a significant burden. This system introduces a semi-automated aggregation technology that leverages generative AI to interactively and quickly consolidate Gen-Ba knowledge fragments that can be collaboratively symbol-grounded in workshops (C).

**(B)    Loose Gen-Ba Knowledge Combination**

There is a need for technology that loosely links Gen-Ba knowledge fragments, which are captured in the Gen-Ba through (A), with documents such as manuals of field operations. This loose structuring of Gen-Ba knowledge is necessary for utilization in internalization workshops (C). Here, we extend the procedure-based and purpose-based knowledge graph (Ijuin et al., 2023) (Inoue et al.,2023), which has been developed and proven effective in caregiving and maintenance inspection fields. By leveraging generative AI technology, we are developing an interactive method to link field work processes and objects with Gen-Ba knowledge fragments using the knowledge graph. Loosely structured Gen-Ba knowledge fragments will be accumulated in the digital fragmented knowledge database.

**(C)    Automatic Recommendation in Internalization**

The system supports efficient and effective collaborative symbol grounding (internalization) in workshops involving both skilled and unskilled participants, utilizing loosely structured digital fragmented knowledge and generative AI. Specifically, we develop (1) automatic classification technology for Gen-Ba knowledge to make workshops more efficient (Ogawa et al., 2024), and (2) recommendation technology for Gen-Ba knowledge to make workshops more effective and stimulate discussions (Ogawa et al., 2025).

While this Digital Knowledge Twin System has been partially developed, the full integration and evaluation of generative AI remain tasks for the future.

**DISCUSSION**

Knowledge management systems that leveraged AI in the twentieth century required knowledge to be described in a fully formalized manner — as logical expressions or IF-THEN rules — at the time of registration, which demanded enormous effort in knowledge preparation. Furthermore, these systems faced the critical limitation that the knowledge needed at the time of use could not always be successfully retrieved, ultimately resulting in systems that were rarely put to practical use. In contrast, the essential characteristic of generative AI lies in its ability to make use of even incomplete or fragmented knowledge that has not been fully formalized.

The GenAI SECI model proposed in this paper is distinguished by its introduction of "Digital Fragmented Knowledge" — a body of partial and fragmented knowledge — as a new concept, while preserving the four knowledge conversion processes of the original SECI model, and by leveraging the capabilities of generative AI to collect, integrate, and utilize such knowledge. The position that existing models require revision to adequately address knowledge management with generative AI is shared with the GRAI framework of Böhm & Durst (2026) and the AKI model of Kirchner & Scarso (2026). However, a fundamental distinction of the proposed model from both is that generative AI is positioned not as a new actor or agent, but strictly as an auxiliary means to support human knowledge creation (Table 1).

The AKI model introduced "Artificial Knowledge" as the type of knowledge handled by generative AI. "Artificial Knowledge," generated by AI in cyberspace and refined through contextual understanding in dialogue with humans, shares certain characteristics with "Digital Fragmented Knowledge" in the present model. Nevertheless, there is an important distinction between the two. Whereas "Artificial Knowledge" is presented to humans in an organized, explicit form, what is extracted from "Digital Fragmented Knowledge" and presented to humans consists of "knowledge fragments" — which may or may not take the form of explicit knowledge — and their internalization must be carried out proactively and reflectively by the human gaining insights from "knowledge fragments".

For knowledge to become practical, action-oriented understanding grounded in the actual workplace, it is essential that humans undergo a process of genuine comprehension — what might be described as knowledge "clicking into place" — by integrating new knowledge with their own experiential understanding. For this reason, the GenAI SECI model places particular emphasis on workshops, as illustrated through the case of the Digital Knowledge Twin System. This internalization process corresponds to "Reflective Observation" in Kolb's (1984) experiential learning model. Moreover, by incorporating not only one's own experiences and self-recorded "knowledge fragments" but also the "knowledge fragments" of colleagues in the same workplace, it is conceived that generative AI-augmented individual reflective observation, or collaborative reflective observation among organizational members through workshops, can be realized.

As prior research has noted, knowledge generated by generative AI is not, in itself, symbol-grounded. It is the human being who performs symbol grounding. The most distinctive and defining feature of the GenAI SECI model is precisely this: the use of generative AI to facilitate and support human-led symbol grounding — that is, the internalization of knowledge by the human subject.

**Table 1.** Comparison of GRAI, AKI, and GenAI SECI models

|  | GRAI model<br>Böhm & Durst (2025) | AKI model<br>Kirchner & Scarso (2026) | GenAI SECI model<br>This paper |
|---|---|---|---|
| **Model foundation** | Extension of the SECI model. Retains the original four conversion modes while adding an agent layer. | New model inspired by the SECI model. Preserves the knowledge conversion framework. | Inherits the four SECI conversion processes. Positioned as an expansion to the existing model. |
| **Role of GenAI** | New actor | New actor | Auxiliary means |
| **Types of knowledge** | Tacit, Explicit Knowledge | Tacit, Explicit, Artificial Knowledge | Tacit, Explicit, Digital Fragmented Knowledge |
| **Approach to internalization** | Promotes internalization through AI-driven learning support, personalization, and role-play simulations. | Human learning from AI within Mentoring Ba. Humans absorb explicit knowledge provided by AI (Absorption mode). | Corresponds to Reflective Observation in Kolb's (1984) experiential learning model. Emphasizes workshop-based experiential understanding. |
| **Symbol grounding** | No explicit mention. | Implies that Artificial Knowledge is not symbol-grounded, but offers no concrete countermeasures. | Explicitly states that symbol grounding is performed by humans. GenAI is positioned as a tool to support this process. |
| **Model complexity** | Moderate: Conceptually straightforward extension via the addition of an agent layer. | High: 9 conversion modes and 9 Ba. Loss of parsimony is acknowledged as a limitation. | Simple: Retains the original 4 conversion modes. Prioritizes ease of implementation and practical applicability. |

**CONCLUSION**

This paper addressed a critical challenge in the field of knowledge management: the immaturity of systematic knowledge management models that handle both tacit and explicit knowledge in an integrated manner in the context of generative AI utilization. While generative AI has rapidly expanded the possibilities for managing explicit knowledge, the fragmentary and partial nature of workplace tacit and latent knowledge has remained largely unaddressed by existing theoretical frameworks. To bridge this gap, we proposed the "GenAI SECI model" as an updated version of the SECI model, redesigned around the capabilities of generative AI.

The central contribution of the GenAI SECI model is the introduction of "Digital Fragmented Knowledge" as a new category of knowledge. Unlike conventional knowledge management systems that required fully formalized, rule-based descriptions, Digital Fragmented Knowledge captures partial and fragmentary field knowledge (knowledge fragments) accumulated in cyberspace. The key insight is that generative AI has made it possible to utilize such incomplete knowledge, which existing knowledge management frameworks previously struggled to handle. In the GenAI SECI model, the four original SECI conversion processes are retained, while generative AI is strategically applied to Externalization, Combination, and Internalization to collect, integrate, and leverage these knowledge fragments.

As prior research has noted, knowledge generated by generative AI is not, in itself, symbol-grounded. Symbol grounding must be performed by human beings through their own lived experience and reflection. The most defining feature of the GenAI SECI model is that it uses generative AI precisely to facilitate and support this human-led symbol grounding — that is, the internalization of knowledge fragments into genuine tacit understanding. This represents a point of departure from preceding models and constitutes the originality and novelty of the proposed approach.

To demonstrate the practical applicability of the proposed model, this paper also presented the Digital Knowledge Twin System, currently under development, showing that the model is not merely conceptual but can be realized as a concrete system architecture.

At the same time, the GenAI SECI model remains an early-stage proposal, and a number of challenges remain to be addressed. First, empirical validation across diverse workplace settings — including manufacturing, healthcare, agriculture, and maintenance — is needed to assess the model's generalizability and to refine its theoretical foundations. Second, further conceptual development of Digital Fragmented Knowledge is required. Finally, as generative AI capabilities continue to evolve rapidly, it will be essential to continuously reassess the model's assumptions and scope in order to ensure that it remains a relevant and robust framework for knowledge management practice.